\documentclass[10pt,twocolumn,letterpaper]{article}
\pdfoutput=1
\usepackage{cvpr}
\usepackage{times}
\usepackage{epsfig}
\usepackage{graphicx}
\usepackage{amsmath}
\usepackage{amssymb}

\newcommand*\samethanks[1][\value{footnote}]{\footnotemark[#1]}

\usepackage{mathtools}

\DeclarePairedDelimiter\floor{\lfloor}{\rfloor}

\cvprfinalcopy 

\def\cvprPaperID{****} 

\begin{document}

\title{Semi-Supervised Learning for Cancer Detection of Lymph Node Metastases}

\author{Amit Kumar Jaiswal$\thanks{These authors are the corresponding authors and contribute equally to this study}$\\
University of Bedfordshire\\
{\tt\small amitkumar.jaiswal@beds.ac.uk}
\and
Ivan Panshin$\samethanks$\\
Perm State University\\
{\tt\small ivan.panshin@protonmail.com}
\and
Dimitrij Shulkin$\samethanks$\\
Schaeffler Group\\
{\tt\small dimitrijshulkin@gmail.com}
\and
Nagender Aneja\\
Universiti Brunei Darussalam\\
{\tt\small nagender.aneja@ubd.edu.bn}
\and
Samuel Abramov\\
Abramov Software GmbH \& Co. KG\\
{\tt\small abramov@abramov-software.de}
}

\maketitle

\begin{abstract}
Pathologists find tedious to examine the status of the sentinel lymph node on a large number of pathological scans. The examination process of such lymph node which encompasses metastasized cancer cells is histopathologically organized. However, the task of finding metastatic tissues is gradual which is often challenging. In this work, we present our deep convolutional neural network based model validated on PatchCamelyon (PCam) benchmark dataset for fundamental machine learning research in histopathology diagnosis. We find that our proposed model trained with a semi-supervised learning approach by using pseudo labels on PCam-level significantly leads to better performances to strong CNN baseline on the AUC metric. 
\end{abstract}

\section{Introduction}
The scale of digitization of pathology scans is still moderate, and recent research has proclaimed heterogeneity and disagreement amongst pathologists’ diagnoses \cite{topol2019high}. However, high-resolution digitization of microscopic images has inspired computer vision researchers to work in the field of pathology diagnosis. The digitization of whole-slide images (WSI) from glass slides has stimulated researchers to implement state-of-the-art convolutional neural networks (CNNs) in medical imaging. A CNN trained on patches extracted from WSI serves to recognize metastatic cancer detection. CNNs has been exhibited to perform better than pathologists in several tasks. This is partly due to the success of ImageNet 2012 challenge\footnote{http://www.image-net.org/challenges/LSVRC/2012/} and also due to the adaptability of CNNs to medical imaging applications. CNNs comprising of different layers of nodes are essentially pattern recognizers. This property of CNNs has been exploited in medical imaging. A CNN trained on a set of images that have been split into patches correctly labeled by well qualified and medical practitioners can differentiate different parts of an image. A trained CNN network can accept an input of unlabeled image to predict if there is a cancer tumor cell or not. Some studies have compared the performance of pathologists with algorithmic results showing some algorithms performed better in terms of accuracy and time efficiency \cite{bejnordi2017diagnostic,golden2017deep}. Liu et al. \cite{liu2017detecting} implemented CNN on Camelyon16\footnote{https://camelyon16.grand-challenge.org/Data/} dataset for lesion-level tumor detection and achieved above 97\% AUC score in comparison to 73.2\% sensitivity achieved by a human pathologist. Additionally, the approach found that two slides in the training set erroneously labeled.

Further, in recent years deep convolutional neural networks (DCNNs) have improved significantly in the area of computer vision including image recognition and have been widely applied and accepted to enhance healthcare facilities. Litjens et al.~\cite{litjens2017survey} classified digital pathology and microscopy techniques in three broad categories namely (1) detection, segmentation, and classification of nuclei, (2) segmentation of a large organ, and (3) detection and classification of a disease. Computerized digital pathology techniques have improved due to the introduction of challenges in pathology. Annotated whole-slide images provided in Camelyon16 challenge allowed participants to use deep learning models such as VGG, ResNet, and GoogLeNet. Top solutions used one of these architectures.

In this work, we present a semi-supervised learning approach that outperforms, even more, the performance of CNN~\cite{veeling2018rotation} in terms of the AUC metric. Our proposed DenseNet based model is evaluated on a slightly modified version of the PCam dataset. The original PCam dataset contains duplicate images due to its probabilistic sampling, however, our evaluation follows the same dataset with no duplicates in it. Otherwise, the same data and splits as the PCam benchmark dataset are maintained.

Our paper is organized into six sections. Having introduced the extent of the paper in Section 1 followed by semi-supervised learning approach in Section 2 which includes the problem formulation, model architecture and the algorithm applied in effectuating the training steps for better performance of the prediction result (the tumor labels). Next, we will discuss the adapted new distribution on PatchCamelyon benchmark dataset including the techniques applied during training steps. In Section 4, we will report the evaluation results followed by related work in Section 5, while Section 6 concludes the paper. 

\section{Semi-Supervised Learning}
We use a semi-supervised learning approach for incremental training of our proposed model to leverage the unlabeled instances for achieving learning performance. Below we formulate the problem and describe our algorithmic approach for detecting metastatic cancer.

\subsection{Problem Settings}
The cancer detection task is a binary image classification problem, where the input is a small (96 x 96px) digital histopathology image \textit{I} and the output is a binary label $l\in \{0,1\}$ stipulating the absence or presence of metastases in small image patches respectively. 

Every single sample in the training set, we optimize the binary cross entropy loss\footnote{https://keras.io/losses/\#binary\_crossentropy}
\begin{align*}
B_{L}(I,l) = -l\log p(Y=1|I) - (1-l)\log p(Y=0|I)
\end{align*}
where $\textit{p}(Y=i|I)$ refers to the probability that the network specifies to the label i. 

\subsection{Model Architecture}
\begin{figure}[h!]
	\includegraphics[width=\linewidth]{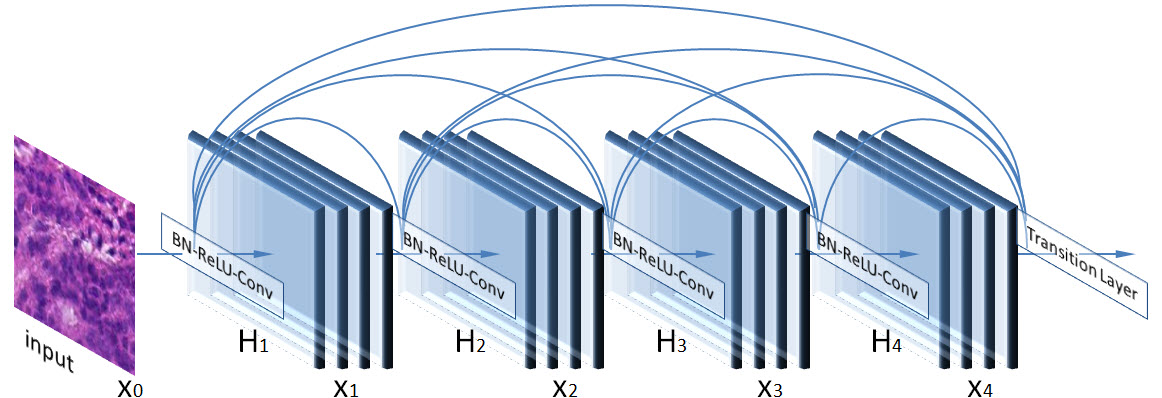}
	\caption{\textbf{DenseNet201 Block Architecture}}
	\label{fig:densenet201}
\end{figure}

For this identification task, we use DenseNet which is a classic CNN architecture that was created~\cite{huang2017densely} in order to solve the vanishing gradient problem~\cite{pascanu2013difficulty}. Unlike other architectures that address this issue, like ResNets~\cite{he2016deep} or highway networks~\cite{srivastava2015training}, whereas in DenseNet all layers are connected so that the information flow between layers in the network is maximal (Figure \ref{fig:densenet201}). In other words, such connectivity pattern introduces $\frac{L(L+1)}{2}$ connections in an L-layer network. Figure 1 illustrates this layout schematically.

To be more precise, the proposed DenseNet201 model uses compression of 0.5 with no bottleneck layers. In other words, if a dense block contains \textit{m} feature-maps, the following transition layer generates $\floor{0.5m}$ output feature-maps.

Moreover, after removing the top layers and instead of fully connected layers, we concatenated the global average pooling (GAP) and global max pooling (GMP) layers including batch normalization (BN) layer. Also, we use dropout layer (0.6) with a dense layer having one output which includes sigmoid activation. 

We concatenated the GAP and GMP layers to use as a slight modification of a strategy described in~\cite{lin2013network}. In this paper, it is proposed to replace the traditional, fully interconnected layers in CNN by GAP. The idea is to create a feature map for each corresponding category of the classification task. Instead of placing fully connected layers over the feature maps, one should take the mean and max of each feature map, and the resulting vector is fed directly after BN and dropout layers into the sigmoid plane. One advantage of GAP and GMP layers across the fully interconnected layers is that it is more native to the convolutional structure by forcing correspondences between feature maps and categories. Another advantage is that there is no parameter for optimization in GAP and GMP layers, which avoids overfitting at this level.

By placing the dropout layer after BN the strategy described in~\cite{li2018understanding} was followed. Past work~\cite{srivastava2014dropout} introduced dropout as an easy way to prevent CNNs from overfitting. It has been proven significantly effective in a variety of machine learning areas such as image classification~\cite{szegedy2015going}. Before the birth of BN, it became a necessity for almost all modern networks and successfully increased their performance against overlay risks despite their amazing simplicity. Past work~\cite{ioffe2015batch} demonstrated BN, a powerful capability that not only accelerated all modern architectures but also improved their strong baselines through their role as regularizers. Therefore, earlier work has employed BN in almost all current network structures~\cite{szegedy2015going,howard2017mobilenets,zhang2018shufflenet} and proves its high practicability and effectiveness.

\subsection{One Cycle Policy}
In this work, we use one cycle policy approach. It was first introduced for SGD~\cite{smith2018super}. One cycle policy is a slight modification of cyclical learning rate policy (CLR) where a minimum and maximum learning rate limits with a step size was specified~\cite{smith2017cyclical}. This policy allows the loss to plateau before the training ends. It combines the advantages of curriculum learning~\cite{bengio2009curriculum} and simulated annealing~\cite{aarts1988simulated}, both of which have a long history of use in deep learning.

\begin{figure}[h!]
	\includegraphics[width=\linewidth]{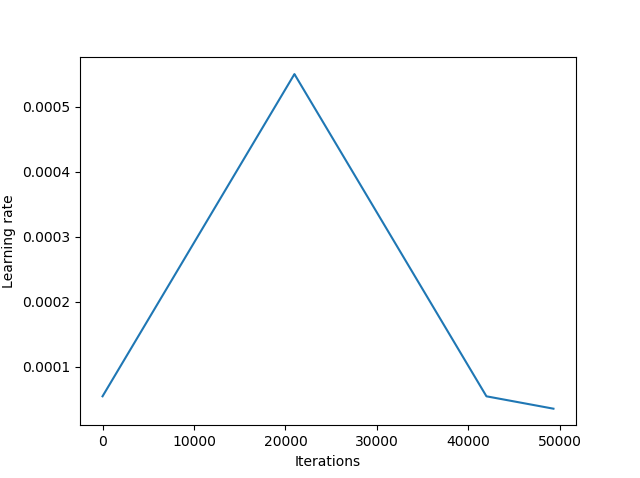}
	\caption{\textbf{One Cyclic Policy - Learning Rate}}
	\label{fig:OneCyclePolicyLR}
\end{figure}

As shown in Figure \ref{fig:OneCyclePolicyLR} the step size is the number of iterations used for each step, and a cycle consists of two such steps - one in which the learning rate (LR) increases and the other in which it decreases. With one cycle policy, the cycle is always smaller than the total number of iterations where the learning rate descends several orders of magnitude less than the initial learning rate for the remaining iterations. 

The maximum learning rate of 0.00055 led to the best results. For selecting the minimum learning rate, we divided the maximum learning rate by a factor of 10.

Momentum and learning rate are closely related. The optimal learning rate depends on the momentum and the momentum depends on the learning rate~\cite{smith2018disciplined}. Also, they found in their experiments that cyclical momentum led to better results. In practice, they recommend choosing two values such as 0.85 and 0.95 and reducing them from the higher to the lower value when the learning rate increases, then returning to the higher momentum when the learning rate decreases.

\subsection{Pseudo Labels in Cancer Identification}
In this section, a semi-supervised learning approach as described in~\cite{lee2013pseudo} is applied where we train a convolutional neural network several times in a supervised manner with labeled and unlabeled data simultaneously. For unlabeled data, pseudo-labels that include the class with the maximum predicted probability are used as if they were real labels.

This method is actually equivalent to entropy regularization~\cite{grandvalet2005semi} where the conditional entropy of class probabilities can be used for a measure of class overlap. By minimizing entropy for unlabeled data, the overlap of the class probability distribution can be reduced. It promotes differentiation between low-density classes, which is often assumed in semi-supervised learning.

Considering multi-layer neural networks with $M$ layers of hidden units it generally follows for the output unit $h_i$ of $k$th layer:

\begin{align*}
h_i^k = s^k\left(\sum_{j=1}^{d^k}W_{i_j}^k h_j^{k-1}+b^k\right), k=1,...,M+1
\end{align*}

where $s^k$ is a non-linear activation function of the $k$th layer, $W_{i_j}^k$ is the weight of $k$th layer connecting input unit $j$ with output unit $i$, $h_j^{k-1}$ is the input value of previous layer, $b_i^k$ is the  bias factor of $k$th layer corresponding to output unit $i$, $f_i=h_i^{M+1}$ are output units used for prediction of target class and $x_j=h_j^0$ are input values. Since this is a binary classification problem, the sigmoid function is used for the output representing probability of true positive label.

The global network is to be trained by minimizing supervised loss function:
\begin{align*}
\sum_{i=1}^{C}L(y_i,f_i(x))
\end{align*}

where $C$ is the number of labels, $y_i$ is the 1-of-K code of the label, $f_i$ is the network output for $i$th label and $x$ is the input vector. Since sigmoid is used for the output for the cancer classification task, cross entropy is given as:
\begin{align*}
L(y_i,f_i(x))=-y_ilogf_i-(1-y_i)log(1-f_i)
\end{align*}

Incorporating pseudo labels into overall loss function gives the following expression:
\begin{align*}
L=\frac{1}{n}\sum_{m=1}^{n}\sum_{i=1}^{C}L(y_{i}^m,f_{i}^m)+\alpha(t)\frac{1}{n'}\sum_{m=1}^{n'}\sum_{i=1}^{C}L({y'_{i}}^m,{f'_{i}}^m)
\end{align*}
where $n$ is the number of mini-batch in labeled data for stochastic gradient descent (SGD), $n'$ for unlabeled data, ${f_{i}}^m$ is the output units of $m$'s sample in labeled data, $y_{i}^m$ is the label of that, ${f'_{i}}^m$ for unlabeled data, ${y'_{i}}^m$ is the pseudo-label of that for unlabeled data and $\alpha(t)$ is a coefficient balancing them.

Entropy regularization~\cite{grandvalet2005semi} is a way to benefit from unlabeled data within the maximum a posteriori estimate. This scheme allows separating low-density classes without modeling the density by minimizing the conditional entropy of class probabilities for unlabeled data:
\begin{align*}
H(y|x') = \sum_{m=1}^{n'}\sum_{i=1}^{C}P(y_i^m=1|x'^m)\log P(y_i^m=1|x'^m)
\end{align*}
where $n'$ is the number of unlabeled data, $C$ is the number of classes, $y_i^m$ is the unknown label of the $m$th unlabeled sample and, $x'^m$ is the input vector of $m$th unlabeled sample. The entropy is a measure of class overlap. As class overlap decreases, it lowers the density of data points at the decision boundary. The mean average precision (MAP) estimate is defined as the maximizer of the posterior distribution:

\begin{align*}
C(\Theta,\lambda) = \sum_{m=1}^{n} \log P((y^{m}|x^{m};\Theta) - \lambda H(y|x';\Theta))
\end{align*}

where $n$ is the number of labeled data, $x^m$ is the $m$th labeled sample, $\lambda$ is a coefficient balancing two terms. By maximizing the conditional log-likelihood of labeled data (the first term) with minimizing the entropy of unlabeled data (the second term), one can get the better performance using unlabeled data.

Pseudo-labels are target classes for unlabeled data as if they were real labels. For the first training run, only the labeled data was used. From the second fine-tuning training run, the following assumption was made:
\begin{align*}
y'_i  = \begin{cases}
1     & \text{ if } P(TP) > 0.9 \\
0   & \text{ if } P(TN) < 0.1 \\
\end{cases}
\end{align*}

where $P(TP)$ is the predicted probability of a true positive label and $P(TN)$ is the predicted probability of a true negative label. With this assumption, fine-tuning training runs were repeated five times. After each fine-tuning training run, it is possible to increase the pseudo label set until a certain convergence is achieved. 

Because the total number of labeled data and unlabeled data is quite different and the training balance between them is quite important for the network performance, pseudo labels with the ratio of 1:1 to the training and validation set were added considering the balance between assumed true positive and true negative pseudo labels. With this approach, it was possible to increase the area under the curve (AUC) of DenseNet201 model after ten fine-tuning training runs.

\section{Adapting New PatchCamelyon Data}
We use the \textit{PatchCamelyon}, a comprehensive patch-level data set derived from Camelyon16 data. In this context, a new benchmark is developed that can accommodate the high volume, quality, and diversity of Camelyon16. The PCam dataset contains 327680 patches extracted from Camelyon16 at a size of 96 x 96 pixels with 10x magnification, selected using a hard negative mining regime. Since metrics at slide-level potentially obscure the relative performance of patch-level models. It has been proposed earlier~\cite{veeling2018rotation} to validate them on a patch-level task. Through this dataset, the task of histopathology diagnosis becomes accessible as a challenging benchmark for fundamental machine learning research. Based on the PCam dataset presents results that are consistently better than results of Camelyon16 state-of-the-art approaches, including~\cite{liu2017detecting,wang2016deep}.

We analyse the 49\% of test data in the first phase, we found no issue with the distribution of various histopathological scans. As the training data is huge with labeled and unlabeled data. However, simply training a neural network on the PCam dataset to predict labels turns out yielding very poor results. To address this issue, proper distribution of targets in a test data is needed and we employed a slightly modified version\footnote{https://www.kaggle.com/c/histopathologic-cancer-detection/data} of the original PCam dataset for this work.

\subsection{Training}
For cancer detection task, we trained our model for ten times with new re-prediction of pseudo labels after each training run, where each training run consists of seven epochs. Before training, we removed 498 images with too many white and black pixels which contain no structure information as outliers from the training set to reduce noise which is illustrated in Figure 3.

\begin{figure}[h!]
	\includegraphics[width=\linewidth]{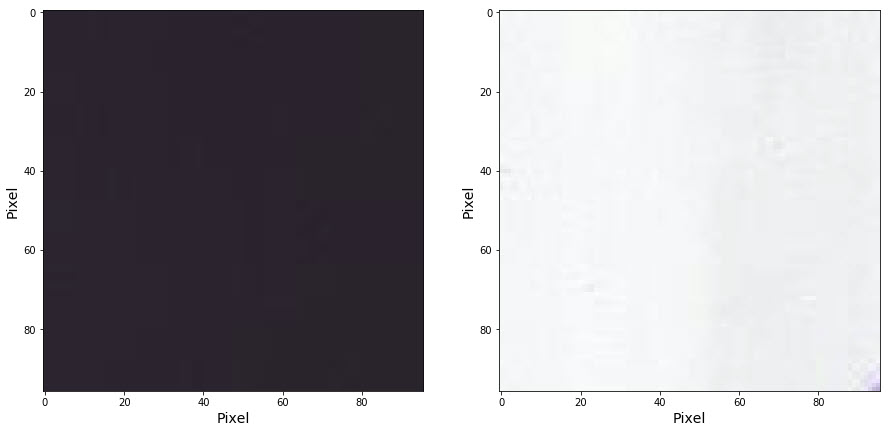}
	\caption{\textbf{Images as Outliers in the Train Set}}
\end{figure}

Finally, we resize the images from 96 x 96 to 224 x 224 pixel as the pre-trained models were originally trained on this size. After each semi-supervised learning run, more and more pseudo labels could be predicted, thus the training corpus could be increased where we perform random split to train and validation set.

Moreover, we apply a set of 10 online data augmentations. We describe the particular transformations in Section 3.2.

\subsection{Test Time Augmentation}
We apply test-time augmentation (TTA) during testing. TTA is a powerful technique that refers to performing data augmentation on a test image in order to get several versions of it and average predictions for them. 

Test-time augmentation has been shown to improve the performances of computer vision algorithms~\cite{wang2018automatic}. Typically transformations include flipping, cropping, rotating, scaling, etc. Having applied a set of transformations during test-time include ten transformations: horizontal flip, vertical flip, rotation from -45 to +45 degrees, cropping each side by 0-20\%, scaling by 80-120\%, translation from -20\% to +20\% (per axis), sharpening and overlying the results with the original using an alpha between 0.0 and 1.0, embossing and overlying the results with the original using an alpha between 0.0 and 1.0, Gaussian noise, changing hue and saturation. In other words, for each original image in test set there are ten modified versions. The model makes eleven predictions and these predictions are blended with equal weights giving the final prediction for the image. 

The transformations for TTA are identical to the ones used in cross-validation. This is done so that cross-validation can be used as a reliable metric of how well an algorithm performs on unseen data (test set).

\section{Evaluation}
\subsection{Results}
We apply semi-supervised learning approach on different pre-trained models\footnote{We experimented with all other models as mentioned and comparing it in Table 1} such as VGG16, InceptionResNetV2, InceptionV3, Xception, ResNet101 and DenseNet201 in the sense of transfer learning.

\begin{table}[!h]
	\caption{\textbf{Evaluation Results}} 
	\label{table:comparison1}
	\centering 
	\resizebox{\linewidth}{!}{%
		\begin{tabular}{|c|c|c|c|} 
			\hline\hline 
			\textbf{Model}  & \textbf{51\% Test Data} & \textbf{49\% Test Data} & \textbf{100\% Test Data}  \\ 
			\hline
			VGG16             & 0.9768 & 0.9721 & 0.9745  \\
			\hline
			InceptionResNetV2 & 0.9764 & 0.9769 & 0.9766  \\ 
			\hline
			Xception          & 0.9748 & 0.9756 & 0.9752  \\ 
			\hline
			InceptionV3       & 0.9758 & 0.9790 & 0.9774  \\ 
			\hline
			SE-ResNet101      & 0.9784 & 0.9781 & 0.9783  \\ 
			\hline
			\textbf{DenseNet201} & \textbf{0.9786} & \textbf{0.9802} & \textbf{0.9794}  \\ 
			\hline   
			GDenseNet~\cite{veeling2018rotation} & & & 0.9630 \\
			\hline
	\end{tabular}}
\end{table}

\begin{figure*}
	\minipage{0.5\textwidth}
	\includegraphics[width=\linewidth]{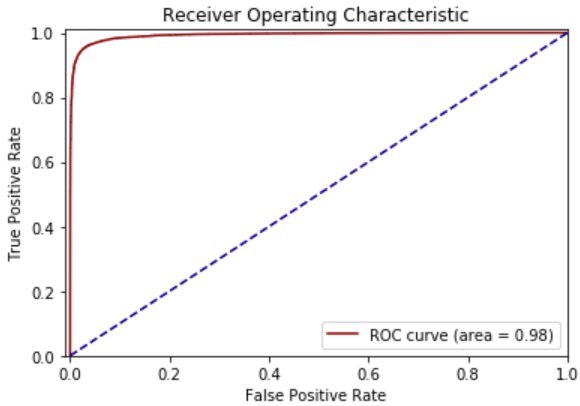}
	\caption{\textbf{Area under the ROC Curve}}
	\label{fig:roc}
	\endminipage\hfill
	\minipage{0.5\textwidth}
	\includegraphics[width=\linewidth]{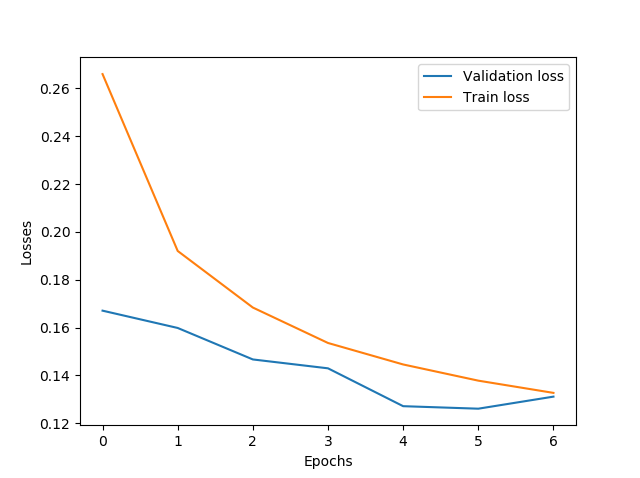}
	\caption{\textbf{Validation and Train Loss}}
	\label{fig:loss}
	\endminipage
\end{figure*}

As shown in the Table \ref{table:comparison1} the DenseNet201 model performs better than other deep CNN models. This is illustrated in \ref{fig:roc}.

The characteristic for all pre-trained models was the fact that they were already over-fitted after 5-7 epochs. The Figure \ref{fig:loss} shows the losses of train and validation sets including pseudo labels during the $10$th fine-tuning run of DenseNet201 model. When considering the validation loss, a sign of overfitting after the 5th epoch can be detected, as the validation loss begins to increase.

\subsection{Ensembles}
Ensemble methods can help to reduce variance~\cite{ju2018relative} and improve the overall performance of machine learning algorithms~\cite{fawaz2019deep,macko2019improving,liu2018make}.

For this cancer detection task, we obtained the best result with the ensembling technique. We train several versions of SE-ResNet101 with an extensive TTA after the predictions from all models by averaging with equal weights. Such approach provided the best and the most robust results. 

The biggest downside of such an approach is that it's computationally expensive. We trained 7 SE-ResNet101 for this ensemble. Each model took 6 hours to train and another hour to get predictions using NVIDIA Tesla P100.

\begin{table}[!h]
	\caption{\textbf{Comparison of the best single model and the ensemble}} 
	\label{table:comparison2}
	\centering 
	\resizebox{\linewidth}{!}{%
		\begin{tabular}{|c|c|c|c|} 
			\hline\hline 
			\textbf{Model}  & \textbf{51\% Test Data} & \textbf{49\% Test Data} & \textbf{100\% Test Data}  \\ 
			\hline
			Ensemble (7 SE-ResNet101) & 0.9810 & 0.9822 & 0.9816  \\
			\hline
			Best single model (DenseNet201) & 0.9786 & 0.9802 & 0.9794  \\ 
			\hline   
			GDenseNet~\cite{veeling2018rotation} & & & 0.9630 \\
			\hline
	\end{tabular}}
\end{table}
We evaluated TTA with a different set of transformations (compared to transformations for the DenseNet model). We use 15 transformations: vertical flip, horizontal flip, rotations by 90, 180, 270 degrees, horizontal flip and rotation by 90 degrees, horizontal flip and rotation by 270 degrees, changing brightness, contrast, saturation, hue, and also 3 different combinations of changing brightness, contrast, saturation, and hue. 

As shown in Table \ref{table:comparison2}, the ensemble technique led to the AUC of 0.9816 (evaluated on 100\% of test data) outperforming the best single model as well as the benchmark solution presented in~\cite{veeling2018rotation}.

\section{Related Work}
Veeling et al.~\cite{veeling2018rotation} proposed rotation equivariant CNNs showing that rotation equivariance improved tumor detection on a challenging lymph node metastases dataset. The authors suggested a fully-convolutional patch-classification model that is equivariant to 90" rotations and reflection. The model has shown a notable advance on the Camelyon16 benchmark~\cite{bejnordi2017diagnostic} dataset.

Bejnordi et al.~\cite{bejnordi2017diagnostic} assessed the performance of automated deep learning algorithms at identifying metastases in hematoxylin and eosin–stained tissue regions of lymph nodes of women with breast cancer and compared it with pathologists’ diagnoses in a diagnostic setting. The experiments results revealed that some deep learning algorithms succeeded more excellent diagnostic performance than a panel of 11 pathologists competing in a simulation study intended to mimic regular pathology workflow; algorithm performance was comparable with a specialist pathologist interpreting whole-slide images without time constraints.

Gorelick et al.~\cite{gorelick2013prostate} implemented a two-stage AdaBoost-based classification for automatic prostate cancer detection and grading on hematoxylin and eosin-stained tissue images. The first stage named tissue component classification includes automatic tessellation of an image into superpixels utilizing a graph-cut based approach; extraction of superpixel appearance, morphometric and geometric features; and classification of superpixels in nine tissue component types based on the extracted features using modest AdaBoost. In the second stage, the authors classified cancer versus non-cancer and low-grade versus high-grade cancer utilizing tissue component labeling. The approach produced a 60-times reduction in data size and thus increasing processing efficiency — the results have shown 90\% accuracy for cancer versus non-cancer and 85\% for high-grade versus low-grade classification. The false-negative rate was 12\% for cancer detection and 5\% for high-grade cancer detection.

Sun et al.~\cite{SunComputer2016} implemented deep learning algorithms for lung cancer diagnosis on lung image database consortium (LIDC) database. The authors implemented a convolutional neural network, deep-belief network (DBN), stacked denoising autoencoder (SDAE). CNN architecture comprises eight hidden layers with odd-numbered convolutional layer and even-numbered pooling and sub-sampling. Each convolutional layer employed 12, 8, 6 feature maps and connected to pooling layers with the 5 x 5 kernel. The architecture of DBN was obtained by training and stacking four layers with each layer holding 100 restricted Boltzmann machine (RBM). The architecture of the SDAE model incorporates three layers SDAE with each autoencoder stacked on the top of each other and each autoencoder having 2000, 1000, and 400 hidden neurons with corruption level of 0.5. The highest accuracy of 0.8119 was obtained in using DBN.

Nahid et al.~\cite{nahidHisto2018} first used unsupervised clustering and further used the deep neural network models guided by the clustered information to classify the breast cancer images~\cite{spanhol2016dataset} into benign and malignant classes.

Wang et al.~\cite{2016arXiv160605718W} proposed a deep learning-based system to auto-detect metastatic cancer from whole slide images of sentinel lymph nodes in the Camelyon challenge 2016. The authors compared GoogLeNet, AlexNet, VGG16, and FaceNet after pre-processing of excluding white background space. The highest performance was obtained in using GoogLeNet with 40 times magnification

Steiner et al.~\cite{steinerImpact2018} conducted a study utilizing results from deep learning algorithms for the detection of breast cancer metastasis in lymph nodes. The study involved reviewing 70 slides by six pathologists in two modes assisted, and unassisted wherein the deep learning mode was used to outline interesting regions in assisted mode. The study found that algorithm-assisted pathologists demonstrated higher accuracy than either the algorithm or the pathologist alone.
Pang et al.~\cite{pang2018using} proposed multiple magnification feature embedding (MMFE) as image tile prediction encoder and slice feature extractor. The method considered inputs image tiles in three resolution 256, 1024, and 4096 and scales to 256. The authors reported 78.1\% accuracy in case of MMFE (tile results) and 84.6\% accuracy in case of MMFE (features).

\section{Conclusion}
We found that some of the techniques did not improve the performances of the model. These techniques include progressive learning; focal loss~\cite{lin2017focal}; average, geometric, and power weights for ensembles; training models with the center crop of 32 px instead of resized images. 

Also, some of the models show significant improvement in the semi-supervised learning approach. With this approach, without the common k-fold method, the area under the curve (AUC) of a best single model could be increased after ten fine-tuning training runs from  0.971 to 0.9794 (evaluated on 100\% of test data) outperforming the benchmark solution introduced in~\cite{veeling2018rotation}. 

In general, pseudo labeling technique allows the training set to be enlarged without knowing the correct labels, allowing the model to achieve better generalization where entropy regularization~\cite{grandvalet2005semi} is a way to benefit from unlabeled data within the maximum a posteriori estimate. This opens up new possibilities for practical use of the model, the basic idea of which is that the single model could be continuously improved in the backend with unlabeled patches derived from new WSIs, which are uploaded to the frontend by the pathologist. 

\section*{Acknowledgements}
We would like to thank Frank Ihlenburg for his valuable comments and acknowledge Kaggle for availing the dataset for this work.

{\small
\bibliographystyle{ieee_fullname}
\bibliography{egbib}
}

\end{document}